\documentclass[11pt]{article}

\usepackage{acl}

\usepackage{times}
\usepackage{tcolorbox}
\usepackage{pifont}
\usepackage{tikz}
\usepackage[edges]{forest}
\definecolor{hidden-draw}{RGB}{0,0,0}

\usepackage[edges]{forest}
\usepackage{xcolor}

\usepackage[utf8]{inputenc} 
\usepackage[T1]{fontenc}   
\usepackage{hyperref}       
\usepackage{url}
\usepackage{multirow}       
\usepackage{booktabs}  
\usepackage{amssymb}
\usepackage{amsthm}

\usepackage{amsfonts}       
\usepackage{nicefrac}       
\usepackage{tabularx}
\usepackage{longtable}
\usepackage{supertabular}
\usepackage{makecell}
\usepackage{microtype}      
\usepackage{xcolor}         
\usepackage{tabularx}
\usepackage{graphicx}
\graphicspath{ {imgs/} }
\usepackage{svg}
\usepackage{enumitem}
\usepackage{subcaption}
\usepackage{amsthm, amsmath}
\usepackage[linesnumbered,ruled]{algorithm2e}
\usepackage{colortbl}

\definecolor{blue}{RGB}{17,220,247}
\definecolor{purple}{RGB}{163,115,250}
\definecolor{caribbeangreen}{rgb}{0.0, 0.8, 0.6}

\definecolor{GREEN}{RGB}{84,130,53}

\usepackage{wrapfig}

\usepackage{todonotes}

\usepackage{bbm}

\usepackage{algorithm}
\usepackage{algpseudocode}
\SetKwComment{Comment}{$\triangleright$\ }{}

\definecolor{GREEN}{RGB}{84,130,53}

\usepackage{bbm}

\usepackage{anyfontsize}

\usepackage{pgfplots}
\pgfplotsset{compat=1.15}
\usepgfplotslibrary{
  statistics,
  colorbrewer,
  groupplots,
}
\usetikzlibrary{
  patterns,
  shapes.geometric,
  decorations.text,
  matrix,
  fit,
  backgrounds,
  positioning,
}
\tikzset{
  fignode/.style={
    outer sep=0.25em,
  }
}
\tikzset{
  framedfignode/.style={
    outer sep=0.25em,
    inner sep=0.5em,
    rounded corners,
    draw,
  }
}
\colorlet{plotColorNeutral}{gray}
\definecolor{plotColor1}{HTML}{f61a1c}
\definecolor{plotColor2}{HTML}{377eb8}
\definecolor{plotColor3}{HTML}{4daf4a}
\definecolor{plotColor4}{HTML}{984ea3}
\definecolor{plotColor5}{HTML}{FFFFCB}
\definecolor{plotColor6}{HTML}{1e90ff}
\colorlet{plotColorNeutral*}{plotColorNeutral!40}
\colorlet{plotColor1*}{plotColor1!60}
\colorlet{plotColor2*}{plotColor2!60}
\colorlet{plotColor3*}{plotColor3!60}
\colorlet{plotColor4*}{plotColor4!60}
\colorlet{plotColor5*}{plotColor5!60}
\colorlet{plotColor6*}{plotColor6!60}
\pgfplotsset{
    colormap={greenred}{HTML=(4daf4a) HTML=(e41a1c)},
    colormap={redgreen}{HTML=(e41a1c) HTML=(4daf4a)}
}

\theoremstyle{definition}

\title{ Trust but Verify! A Survey on Verification Design for Test-time Scaling }

\author{ Venktesh V \\ Stockholm University \\ \texttt{venktesh.viswanathan} \\\texttt{@dsv.su.se }
\And Mandeep Rathee \\ L3S Research Center \\ \texttt{rathee} \\ \texttt{@l3s.de}  
\And  Avishek Anand \\ TU Delft \\ \texttt{avishek.anand}\\ \texttt{@tudelft.nl}  }

\begin{document}

\maketitle
\begin{abstract}
Test-time scaling (TTS) has emerged as a new frontier for scaling the performance of Large Language Models. In test-time scaling, by using more computational resources during inference, LLMs can improve their reasoning process and task performance. Several approaches have emerged for TTS such as distilling reasoning traces from another model or exploring the vast decoding search space by employing a verifier. The verifiers serve as reward models that help score the candidate outputs from the decoding process to diligently explore the vast solution space and select the best outcome. This paradigm commonly termed has emerged as a superior approach owing to parameter free scaling at inference time and high performance gains. The verifiers could be prompt-based, fine-tuned as a discriminative or generative model to verify process paths, outcomes or both. Despite their widespread adoption, there is no detailed collection, clear categorization and discussion of diverse verification approaches and their training mechanisms. In this survey, we cover the diverse approaches in the literature and present a unified view of verifier training, types and their utility in test-time scaling. Our repository can be found at \url{https://github.com/elixir-research-group/Verifierstesttimescaling.github.io}.




\end{abstract}

\begin{table*}[hbt!]
    \centering
    \vspace{{-2em}}
    \resizebox{1\textwidth}{!}{
    \begin{tabular}{lllp{1cm}}
    \toprule
        \textbf{Category} & \textbf{Approach} & \textbf{Approach Description} & \textbf{Type}  \\
    \midrule
        \multirow{3}{*}{\textbf{Heuristic}}     & (1) Deepseek-R1~\citep{deepseek-r1} & Heuristic check for domain-specific problems &  O \\

        & (2) RvLLM \cite{rvllm_rule_based} & Uses domain specific rules to verify solution candidates & P \\

           \hline
   \multirow{5}{*}{\textbf{Discriminative}}       & (3) Naive ORM~\citep{cobbe_2021_ORM} & Trains solution-level and token-level verifiers on labeled-dataset & O  \\
        & (4)  OVM~\citep{yu-etal-2024-ovm} & Train a value model under outcome supervision for guided decoding & O \\
     
       & (5) Naive ORM ~\citep{chen2024tree} & SFT a domain-specific LM as a discriminator & O \\
        & (6) Naive PRM~\citep{lightman2024lets} & SFT an LM as a PRM on each reasoning step over mathematical tasks & O \\
        & (7) AutoCv \cite{autocv}& Trains an ORM first to use its confidence variations as step-level annotations to train a PRM  & O, P \\
        & (8) Math-Shepherd \cite{wang-etal-2024-math} & Trains discriminative PRM using automated process-level annotations  & P \\
        & (9) DiVeRSe \cite{li-etal-2023-making}  & Employs voting and step-wise voting verifier trained using binary cross-entropy loss & P\\
        & (10) Alphamath \cite{alphamath} & Trains a process-level verifier using data from MCTS rollouts & P \\
        & (11) $Re^2Search++$ \cite{raggym} & Trains a ceritic/verifier with contrastive loss on preference data & O  \\

        & (12) VersaPRM \cite{zeng2025versaprm} & extends to domain beyond math by augmenting MMLU with step-level annotations & P \\
                   \hline

       \textbf{Generative} \\
       - Classical SFT & (13) Generative Verifier~\citep{zhang2024generative} & Exploit the generative ability of LLM-based verifiers via reformulating the verification & O, P \\

& (14) UQ-PRM \cite{ye2025uncertaintyaware} & Quantifies uncertainty of verifier for each step of reasoning & P \\

        & (15)  WoT~\citep{zhang-etal-2024-wrong} & Multi-Perspective Verification on : Assertion, Process, and Result wiht single verifier & O, P\\
        & (16) Multi-Agent Verifiers~\citep{lifshitz2025multiagent} &  Multi-Perspective Verification with multiple verifiers & O, P \\ 

        & (17) SyncPL \cite{SyncPL} & Trains a generative process verifier using criteria based data generation & P\\
        \cmidrule{2-4}
        - Self-verification  & & & \\
        (classical SFT) 
        & (18) ReVISE \cite{lee2025revise} & Trains on self-generated training data using SFT to elicit self-verification  & O \\
        & (19) Self-Reflection Feedback~\citep{li2025learningreasonfeedbacktesttime} & formulate the feedback utilization as an optimization problem and solve during test-time & O \\
        & (20) ToolVerifier \cite{mekala-etal-2024-toolverifier} & Trains a LLM on data generated for verifying tool usage & P \\
        & (21) T1 \cite{t1} & Trains a LLM using SFT by distilling from tool integrated data generation for self-verification & O \\

            
        \cmidrule{2-4}
       - RL-based 
        & (22) GenRM \cite{mahan2024generativerewardmodels} & Trains a generative verifier on preference data by unifying RL from human and AI feedback. & P \\
       
       &(23)  V-STaR~\citep{hosseini2024vstar}  & Verifier trained on both accurate and inaccurate rationale/reasoning chains & P \\

        & (24) VerifierQ \cite{qi2024verifierqenhancingllmtest} & Trains process-level verifier using Q-Learning & P \\

        & (25) PAV \cite{setlur2025rewarding} & Trains based on measure of progress achieved using a step than absolute values & P \\

       & (26)  RL-Tango \cite{zha2025rltangoreinforcinggenerator} & Trains a generator and verifier in tandem using RL & P\\

       & (27)  S2R \cite{ma-etal-2025-s2r} & trains ORM and PRM with RL & O, P \\

\midrule
       {\textbf{Reasoning}} & (28) Heimdall \cite{heimdall_reasoning_verification} & Adopts a LRM and pessimistic approach for verification reducing uncertainty & P\\
        & (29) Think-J \cite{thinkJ} & LRM fine-tuned to serve as LLM judge & P \\

        & (30) ThinkPRM \cite{khalifa2025processrewardmodelsthink} & proposes to train a generative PRM using a reasoning model on synthetic data& P \\
       & (31) J1 \cite{whitehouse2025j1incentivizingthinkingllmasajudge}& Fine-tunes a LRM with RL and classical SFT for verification & O, P\\
        & (32) $RL^V$ \cite{rlv_reasoning} & Trains a LRM on generative verification and RL objectives jointly & P  \\
        & (33) DyVe \cite{zhong2025dyvethinkingfastslow} & Trains a LRM as dynamic verifier through SFT on carefully curated data & P  \\
        & (34) FlexiVe \cite{flexive} & Trains a LRM as dynamic verifier with RL alternating between fast and slow thinking & P  \\
        & (35) OREO \cite{OREO} & Trains a value function by optimizing soft Bellman equation for test-time scaling & P \\
    \midrule


      \textbf{Symbolic}   & (36) Deductive PRM~\citep{deductive_prm} & Generates symbolic programs for process verification by distilling from stronger LLMs  & P \\ 
      & (37) RaLU \cite{li2025reasoningaslogicunitsscalingtesttimereasoning} & Trains a verifier to check alignment of natural language to symbolic program & P \\

      & (38) SymbCOT \cite{symbCOT} & Translates natural language rationales to symbolic outputs for verification using logic deduction & P \\

        & (39) ENVISIONS \cite{ENVISIONS} & Proposes an environemnt guided neuro-symbolic self-training, self-verifying framework & P \\
        & (40) START~\citep{li2025startselftaughtreasonertools} & Self-distilled reasoner with external tool calling for verification & P \\

       &  (41) LMLP \cite{LMLP} &  Learns to generate facts and logic programs over a KB for step-by-step verification & P
        \\

& (42) FOVER \cite{FOVER} &  trains PRMs on step-level labels automatically annotated
by formal verification tools & P
 \\       
 
  
    \bottomrule
    \end{tabular}}
    \vspace{-0.8em}
    \caption{Summary of verifier training mechanisms as applied to test-time scaling. KB- Knowledge Base, O - Outcome, P - Process}
    \label{tab:verification}
\end{table*}

\section{Introduction}
\label{chapter:introduction}

Large language models (LMs) have largely been driven by scaling up train-time compute through large-scale self-supervised pretraining \citep{kaplan2020scalinglawsneurallanguage,hoffmann2022trainingcomputeoptimallargelanguage}.
Recently a new paradigm: \emph{test-time scaling} has been evolving where additional computational resources are allocated during inference, allowing models to further refine their predictions and improve performance.

The central idea behind test-time scaling (TTS) is that increasing the compute budget at inference can yield substantial gains in model performance. This additional compute is used to search over the space of possible solutions. The test-time scaling mechanisms can be divided into two classes namely the \textbf{verifier-free} approaches which generate reasoning traces from larger model and distill them to a smaller LLM or \textbf{verifier-based approaches} which employ an external signal to guide search over solution space to select the best one.

The verification of generated solutions and reasoning paths is central to test-time scaling. Recent studies \cite{setlur2025scalingtesttimecomputeverification} have demonstrated that scaling test-time compute without verifiers is suboptimal and the performance gap between \textit{verification-based} scaling and \textit{verifier-free} scaling widens as the allotted compute increases. 

Common approaches for TTS include letting the LLM generate a long chain-of-thought trajectory \cite{Wei2022ChainOT,Kojima2022ZeroShot,openai2024o1,zeng2024scalingsearchlearningroadmap},
or asking the LLM to iteratively refine the solutions that it has generated \cite{davis2024networksnetworkscomplexityclass,madaan2023selfrefine,du2023improvingfactualityreasoninglanguage,yinetal2023exchange,yinetal2024aggregation} termed as modifying proposal distribution.
Another category of approaches entails sampling multiple candidate solutions 
and then choosing the best one via majority voting \cite{chen2024llm,wang2023selfconsistency,brown2024largelanguagemonkeysscaling,li2024more},
ranking with pairwise comparisons \cite{jiang2023llmblender},
or using an external verifier  or trained reward model\cite{kambhampati2024modulo,stroebl2024inferencescalingflawslimits} \cite{cobbe_2021_ORM,lightman2024lets,zhang2024generative}.

Employing external verifiers or self-verification is crucial for test-time scaling, as they help guide the search process over large reasoning space. \textit{Verification for test-time scaling entails  mechanisms or scoring functions used to evaluate the quality or plausibility of different reasoning paths or solutions from the language model during inference, enabling efficient search or selection among them without access to ground-truth labels.}

The verifiers can be divided into process-based namely Process Reward Models (PRMs) \cite{uesato_PRM,ligthman_PRM} if they verify the LLM reasoning process step-by-step or outcome-based - Outcome Reward Models (ORM) \cite{cobbe_2021_ORM} if they are only concerned about correctness of final solution. However, manual annotation of step-level correctness for training PRMs is time-consuming and infeasible leading to new paradigm of synthetic reasoning path generation and annotation of them using various signals such as final ground truth answer \cite{alphamath}, LLM uncertainty \cite{ye2025uncertaintyaware} or verifier confidence \cite{autocv}.

While a large number of verification approaches have emerged very recently, there is a lack of clear distinction based on how they are trained, how their training data is collected and how they are employed for test-time scaling. 
The diversity in verifier types, synthetic data generation and training objectives for verifiers can also be understood from the lens of \textbf{asymmetry of verification}: where certain tasks are hard to solve but easy to verify or easy to solve but hard to verify \cite{liu2025trustverifyselfverificationapproach,qin2025backtrackbacktracksequentialsearch}.

In this survey, our goal is to present a clear and simple taxonomy for verifiers based on training recipes, objective functions and their inference time utility in test-time scaling. 
Note that while reward models are also used for verification purposes in pre-training, our scope is limited to verification strategies, particularly for test-time scaling. 
We observe based on our survey that the verifier design choices can be ascribed to task-specific challenges due to asymmetry of verification. 
We also present challenges with current approaches and new perspectives on future of verifier training mechanisms. 
We also discuss a training recipe for verifiers that could emerge from recent advances in self-play \cite{zhao2025absolutezeroreinforcedselfplay} as future directions. Details on how we collect literature for this topic can be found at Appendix \ref{app:literature_compilation}

\noindent \textbf{Comparison to adjacent surveys}
While several surveys focus on Test-Time Scaling (TTS) \cite{zhang2025surveytesttimescalinglarge,chung2025revisitingtesttimescalingsurvey}, they do not adequately cover wide-range of verifier types and their training mechanisms in context of test-time scaling. To the best of our knowledge, our survey is the first to do this as verifiers are cornerstone of TTS and a clear taxonomy of approaches help inform their usage for real-world applications. Our repository can be found at \url{https://github.com/elixir-research-group/Verifierstesttimescaling.github.io} and our website can be found at \url{https://elixir-research-group.github.io/Verifierstesttimescaling.github.io/}.
\section{Notions and Concepts in Test-Time Scaling}

\textit{Test-time scaling} refers to allocating additional computational resources to language models (LLMs) during inference to improve their outputs without modifying the model weights. 
Test-time scaling involves three key dimensions: \textit{when verification occurs} (at the outcome or process level), \textit{how exploration is conducted} (in parallel, sequentially, or using hybrid strategies), and \textit{how candidate solutions are selected or refined} (via sampling, reranking, structured search, or self-correction). A clear understanding of these dimensions is critical for designing more robust and capable inference-time reasoning systems.

\subsection{Scaling Paradigms at Inference}
Test-time scaling can be broadly categorized into three paradigms. 

In \textbf{parallel scaling}, the model generates multiple independent outputs simultaneously, often by varying sampling temperature or prompt exemplars to induce diversity \citep{levy-etal-2023-diverse,brown2024largelanguagemonkeysscaling}. These outputs form a candidate set \( \mathcal{S} = \{s_1, \ldots, s_k\} \), from which a selection mechanism \( \mathcal{V} \) identifies the final answer \( s^* = \mathcal{V}(\mathcal{S}) \).

\textbf{Sequential scaling}, in contrast, decomposes a problem into intermediate steps or sub-questions. Each step builds on the previous one, producing a sequence \( \{sq_1, \ldots, sq_T\} \) where each \( sq_t = \text{LLM}(sq_{t-1}, c_t) \) depends on the prior reasoning step and contextual information \( c_t \) \citep{madaan2023selfrefine, du2023improvingfactualityreasoninglanguage}.

\textbf{Hybrid scaling strategies} integrate both approaches. A model may first generate multiple reasoning paths in parallel, then iteratively refine or select from them using sequential reasoning. Notable examples include beam search \citep{OREO,snell_TTS}, tree/graph/forest-of-thoughts \citep{tree_of_thought,graph_of_thoughts,bi2025forestofthought}, and constrained Monte Carlo Tree Search (MCTS) \citep{constrained_MCTS}, where exploration and refinement are tightly coupled.


\subsection{Verification Strategies}
No matter how the candidate solutions are generated, selecting a reliable answer often requires \textit{verification}. \textbf{Outcome verification}, also known as outcome reward modeling (ORM), evaluates whether the final answer is correct without inspecting intermediate steps. These models are relatively easy to train using weak supervision and are commonly used in reranking or filtering \citep{cobbe_2021_ORM,lightman2024lets,zhang2024generative}.

In contrast, \textbf{process verification}—or process reward modeling (PRM)—assesses the reasoning path itself, step by step. This approach typically relies on annotated reasoning traces and provides more fine-grained evaluations of model behavior \citep{uesato_PRM,zhang2025lessonsdevelopingprocessreward,zheng2025processbenchidentifyingprocesserrors}. Recent work also explores \textit{self-verification}, where the model critiques its own reasoning path without external feedback \citep{self_verification_1,lee2025revise,t1}.

\begin{figure}
\centering
{\footnotesize
\definecolor{nodefill}{RGB}{238, 238, 238}
\forestset{qtree/.style={for tree={parent anchor=south, 
           child anchor=north,align=center,inner sep=3pt, fill=nodefill, rounded corners=2pt}}}

\begin{forest} , baseline, qtree
    [Verifier training\\Approaches
        [Heuristic]
         [Generative
            [Self-verify / \\ LLM as Judge \\ (classical SFT)]
            [Reasoning-based\\GRM]
            [Fine-tuned\\GRM [RL-based] [Classical \\ SFT]]
        ]
        [Discriminative]
        [Symbolic]
    ]
\end{forest}
}

\caption{Categorization of verifier training approaches.}
\label{fig:categorization_tree}
\end{figure}
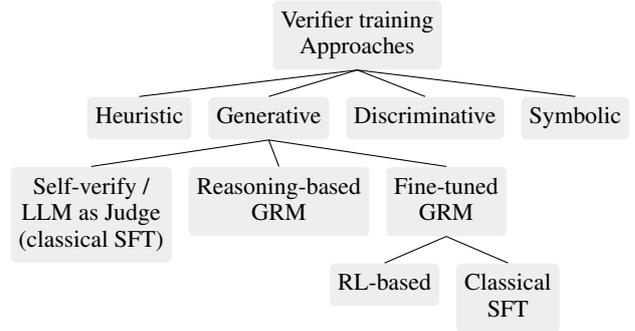

\begin{figure*}
    \centering
    \includegraphics[width=\linewidth]{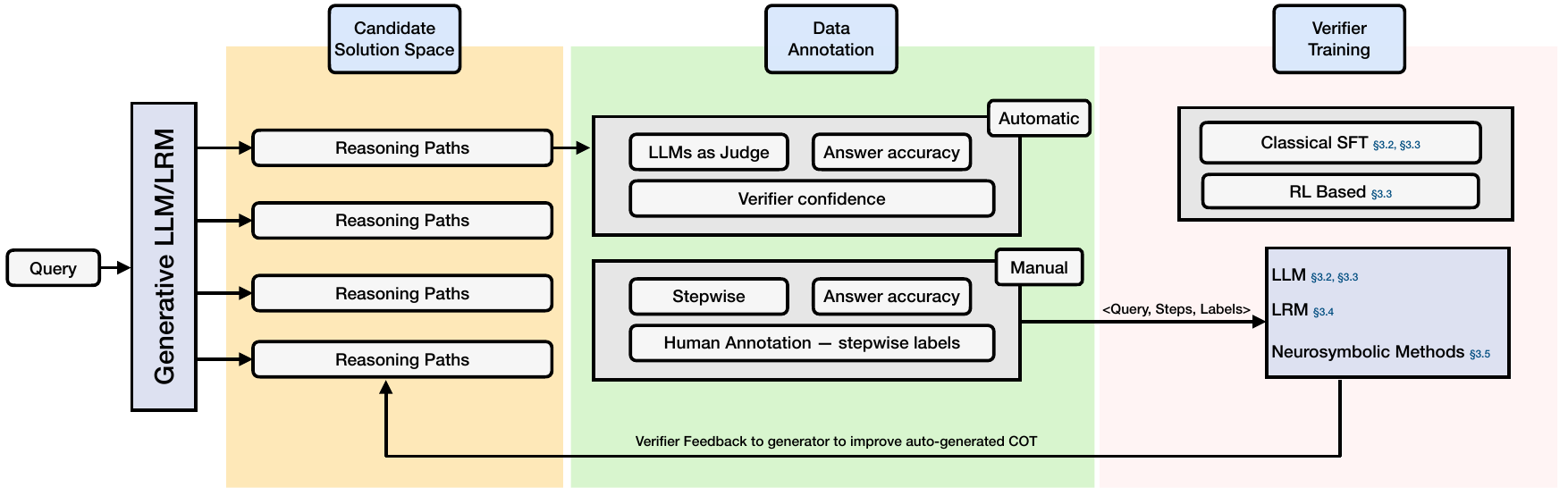}
    \caption{An overview of  verifier training mechanisms}
    \label{fig:unified_framework}
\end{figure*}
 
\section{Training Paradigms and Types of Verifiers}
\label{sec:methods}

We start by giving an overview of the verifier training procedure as in Figure~\ref{fig:unified_framework} that is organized along three key dimensions: (1) the \textit{nature of the supervision} (heuristic, supervised, reinforcement learning), (2) the \textit{type of data used} (manual annotations, synthetic rollouts, symbolic derivations), (3) the \textit{output modality }(discriminative vs. generative).
We also categorize training strategies for verifiers used in test-time scaling in Figure~\ref{fig:categorization_tree}. 
This taxonomy not only provides a conceptual map of existing methods but also considers design trade-offs that affect performance, scalability, and robustness. 
Understanding the diversity and limitations of current training strategies is crucial for choosing the right approach for a given task, whether it be math reasoning, fact verification, or open-domain QA. 

Verifier training approaches vary in how they model solution correctness: via rules, scores, rationales, or logic chains and differ in their data needs and robustness. Discriminative models are lightweight but narrow in domain; generative and reasoning-based verifiers better capture complex verification needs, while symbolic verifiers offer formal guarantees. 


\subsection{Heuristic Verifiers}
Early verification strategies primarily relied on heuristics to verify final solutions such as output fluency, plausibility, adherence to format \cite{deepseek-r1} where LLMs are trained to self-verify using these heuristics as rewards for RL-based training using Group Relative Policy Optimization (GRPO). RvLLM \cite{rvllm_rule_based} designs an Expert Specification language (ESL) that can be used by experts to design domain specific rules for verification. However, rule based verification approaches are not scalable as it is time-consuming to author comprehensive rules. They may also fail to capture semantic variations in expression (for example: ``2 hours" and ``120 minutes" imply the same) resulting in incorrect verification. Hence, model-based verifiers are proposed as covered in subsequent sections.               \vspace{-0.6em}

\subsection{Discriminative Verifiers}

Discriminative verifiers frame the verification task as a supervised classification problem, where models are trained to assign correctness scores to reasoning steps or final answers. These models are commonly used in tasks that are relatively \textbf{easy to verify}, where correctness can be inferred from observable outputs. A standard distinction exists between \textbf{Outcome Reward Models (ORMs)}, which evaluate only the final answer, and \textbf{Process Reward Models (PRMs)}, which assign scores to each step in the reasoning trajectory \cite{cobbe_2021_ORM, ligthman_PRM, yu-etal-2024-ovm}.

PRMs are typically fine-tuned on datasets with binary step-level annotations \cite{uesato_PRM, zhang2025lessonsdevelopingprocessreward}. Given a reasoning step, the model encodes the input and outputs a binary score using a classification head optimized with cross-entropy loss. Final solution quality is often estimated by aggregating the predicted scores across steps \cite{snell_TTS, wu2024scaling}.

However, manually creating step-level labels is labor-intensive, prompting the development of automated annotation strategies. In many approaches, a reasoning step is deemed correct if a valid final answer can be generated by an LLM in a fixed number of sampling trials (e.g., via Monte Carlo rollouts) \cite{wang-etal-2024-math, luo2024improvemathematicalreasoninglanguage, alphamath}. This proxy ties the \textbf{answer accuracy} from future rollouts to the value of the current step. Subsequent steps after the first failure are typically discarded.

To avoid information loss by discarding paths, AutoCV \cite{autocv} proposes a confidence-based annotation strategy. An ORM is trained on final-answer correctness, and its confidence change upon adding a new reasoning step is used to estimate that step’s value. A significant drop in confidence suggests a flawed reasoning step, which helps label steps without explicit human annotations.

Despite empirical success, these models often assume stepwise independence. Li et al.~\cite{li2025process} identify that this independence assumption undermines reward distribution across reasoning trajectories. They reformulate step-level verification as a Markov Decision Process (MDP) and introduce a Q-value ranking objective to model dependencies between steps, enabling more consistent and context-aware feedback.

While early work focused on math and code reasoning, more recent approaches extend discriminative verifiers to knowledge-intensive tasks. \textsc{Re$^2$Search++} \cite{raggym} leverages contrastive learning to distinguish correct intermediate actions from incorrect ones in open-domain QA, offering feedback not only on reasoning correctness but also on auxiliary decisions such as query formulation. VersaPRM \cite{zeng2025versaprm} augments the MMLU benchmark with synthetic CoT traces and counterfactual variants, expanding PRM training to general-domain tasks. This helps improve verifier generalization across diverse topics beyond structured tasks like math.

\subsection{Generative Verifiers}
\label{sec:generative}

Generative verifiers leverage the inherent capabilities of large language models (LLMs), including natural language generation, chain-of-thought (CoT) reasoning \cite{chainofthought}, and instruction-following, to assess the correctness of solutions. Unlike discriminative verifiers, which produce scalar confidence scores, generative verifiers produce textual critiques or judgments. This approach not only aligns with the pretraining objective of LLMs but also enables greater interpretability, compositionality, and adaptability during inference \cite{wang2024guiding, brown2024largelanguagemonkeysscaling}.

While prompting off-the-shelf LLMs as judges~\cite{zhen_llm_judge} offers a straightforward way to tap into these capabilities, it often underperforms compared to fine-tuned models on reasoning benchmarks. To overcome this, recent work has focused on training generative verifiers using synthetic supervision or reinforcement learning. We group these approaches into three main categories: (i) supervised fine-tuning (SFT), (ii) reinforcement learning (RL), and (iii) self-verification. These are summarized in Table~\ref{tab:verification} and Figure~\ref{fig:categorization_tree}. Table \ref{tab:hypothesis} gives an overview of hypothesis behind some of important training recipes for verifiers.

\paragraph{Supervised Fine-Tuning (SFT)}
Early work in generative verification \cite{zhang2024generative} fine-tunes LLMs using next-token prediction to jointly generate critiques and correctness assessments for solution candidates. Here, the verifier is trained on synthetic critiques generated for reasoning paths and final answers. Instead of assigning numeric scores, correctness is inferred from the model’s token probabilities over natural language outputs.

Other approaches rely on synthetic supervision from stronger LLMs or tool-augmented critiques \cite{li2025learningreasonfeedbacktesttime, lee2025revise, t1, mekala-etal-2024-toolverifier}. These are used in \textit{self-verification} pipelines, where models are trained to evaluate their own outputs or those of a fixed generator. These approaches employ Rejection Sampling Fine-tuning (RSFT) where only synthetic trajectories leading to correct solutions are retained for SFT of verifier. SyncPL \cite{SyncPL} alternatively proposes to augment RSFT with criteria based filtering like logical consistency of synthetic solution trajectories.

While simple and scalable, these methods can suffer from poor robustness and susceptibility to \textit{reward hacking} \cite{ye2025uncertaintyaware}, where the verifier learns spurious correlations that inflate reward without genuine correctness.
To address these concerns, ~\citet{ye2025uncertaintyaware} introduce \textbf{COT-entropy}, a generative verifier with uncertainty quantification. Their method explicitly penalizes overconfident predictions by modeling token-level entropy during reasoning, improving reliability in process-level verification tasks.

\paragraph{Reinforcement Learning (RL) Approaches}
Reinforcement learning has emerged as a powerful paradigm for training generative verifiers, offering better alignment with preferences and improved generalization.

V-Star \cite{hosseini2024vstar} builds on the StaR framework \cite{zelikman2022starbootstrappingreasoningreasoning} by sampling both correct and incorrect reasoning traces and deriving preference data from correctness labels. The verifier policy is then optimized using Direct Preference Optimization (DPO), encouraging the generation of faithful reasoning steps and accurate final answers. GenRM \cite{mahan2024generativerewardmodels} follows a similar design, fine-tuning a generative verifier on synthetic preference data using preference-based loss functions.

VerifierQ \cite{qi2024verifierqenhancingllmtest} adopts a different RL objective by training the verifier using Q-learning. Here, the verifier estimates step-level Q-values that reflect the expected reward for continuing from a given reasoning step. Conservative Q-learning is used to avoid overestimation, enabling stable training. This formulation allows parallel ranking of solution candidates and supports hybrid search strategies. Rather than assigning absolute values to steps and ignoring incorrect steps like previous approaches, PAV \cite{setlur2025rewarding} measures progress that can be achieved by including a step enabling more exploration when coupled with RL-based training outperforming VerifierQ and other approaches.

\paragraph{Co-evolution of Generator and Verifier}
Fixed-generator settings introduce a bottleneck: if the verifier is trained on poor-quality samples, it risks overfitting or reward hacking. RL-Tango \cite{zha2025rltangoreinforcinggenerator} addresses this by jointly training a generator-verifier pair using Group Relative Policy Optimization (GRPO). The verifier is trained using outcome-level rewards to predict stepwise correctness, which in turn informs the generator’s updates, closing the feedback loop. This co-evolution leads to improved solution quality and better verifier generalization.
S2R \cite{ma-etal-2025-s2r} builds on this idea by encouraging the generator to self-verify, using both outcome- and process-level feedback to guide training. Like RL-Tango, it avoids manual step-level annotations by relying on downstream rewards.

\vspace{-0.5em}
\subsection{Reasoning-Based Generative Verifiers}

While generative verifiers capitalize on the natural generation capabilities of LLMs, they often fall short when applied to tasks that are inherently ``hard to verify'' — a challenge rooted in the \textit{asymmetry of verification}: some problems are easy to solve but difficult to verify, and vice versa. Addressing this limitation has led to the rise of reasoning-based generative verifiers, which employ long-form reasoning and deliberate critique generation, leveraging recent advances in Large Reasoning Models (LRMs) \cite{thinkJ,khalifa2025processrewardmodelsthink}.

ThinkPRM~\cite{khalifa2025processrewardmodelsthink} is one such model that trains a process verifier (PRM) using LRMs in a generative fashion. It utilizes high-quality synthetic reasoning trajectories, generated via automated step-level labeling, to fine-tune the verifier. 
Despite using less annotated data, ThinkPRM demonstrates strong data efficiency and performance, outperforming models trained on over 100$\times$ more human-labeled data.

Other efforts extend this approach by refining the ``LLM-as-a-judge'' paradigm~\cite{zhen_llm_judge}. 
For example, \cite{whitehouse2025j1incentivizingthinkingllmasajudge,thinkJ} fine-tune LRMs to reason through and rank candidate responses based on verification-specific objectives.
Unlike previous approaches requiring enormous data engineering, OREO~\cite{OREO}, frames verification as a byproduct of reinforcement learning. OREO jointly trains a ``policy'' and a ``value function'' within an offline RL framework using a LRM. The training optimizes a soft Bellman equation to closely approximate optimal value estimates. The resulting value function acts as an implicit verifier during test-time search, similar to verifier strategies such as RL-Tango.

While reasoning-based generative verifiers like ThinkPRM and OREO are powerful, they often do not explicitly address robustness and out-of-domain generalization. The verifier from \cite{heimdall_reasoning_verification} targets these challenges by combining long-form reasoning generation with reinforcement learning on diverse data. To further improve robustness, the paper introduces a \textit{pessimistic verification} strategy, which prefers solutions with lower associated uncertainty.
Despite their strengths, most of these methods allocate a fixed compute budget to all verification instances. However, real-world use cases often involve a mix of easy and complex verifications. Allocating equal compute to all instances may lead to inefficiencies, including \textbf{overthinking}, where unnecessary reasoning is generated for obvious cases \cite{overthinking}.

To address this, DyVe\cite{zhong2025dyvethinkingfastslow} and FlexiVe~\cite{flexive} propose adaptive verification frameworks that tailor compute usage to the complexity of the instance. Drawing inspiration from dual-system theories of cognition \cite{kahneman2011thinking}, they alternate between ``fast thinking'' (direct token-level verification) and ``slow thinking''(long-form deliberation). DyVe trains on both simple and complex reasoning traces from datasets like MathShepherd \cite{wang-etal-2024-math}, with stepwise judgments on whether deeper verification is needed. It fine-tunes an LRM using cross-entropy loss to produce labels for simple tasks and to generate full reasoning paths for complex ones.

In contrast, \textit{FlexiVe} avoids the need for labeled signals about complexity. Instead, it uses \textit{Group Relative Policy Optimization (GRPO)}\cite{deepseek-r1} to train a process verifier. The reward function is composite: one component encourages correct verification, and another penalizes unnecessarily long outputs. This incentivizes the verifier to dynamically switch between fast and slow modes based on task difficulty.
\vspace{-0.7em}
\subsection{Symbolic Verifiers}

Despite the advances in process reward models (PRMs) trained using synthetic reasoning traces, these models often suffer from two key limitations: inaccurate step-level annotations due to reasoning drift or hallucinations, and poor generalization to out-of-distribution (OOD) tasks. Symbolic verification methods aim to address these issues by grounding reasoning in formal representations and structured logic, offering correctness guarantees, and improving robustness. Tasks that lend themselves to formal validation, namely mathematical proofs, logical inference, and code reasoning, are thus considered \textbf{easy to verify} under this paradigm.

In this section, we categorize symbolic verification approaches based on their use of symbolic reasoning at inference time or as a training signal. We distinguish between methods that (1) execute or emulate formal logic systems during inference, (2) translate natural language reasoning into symbolic forms for stepwise validation, and (3) augment training data with symbolic verification feedback.

\paragraph{Symbolic Reasoning at Inference Time}
\textit{SymbCOT} \cite{symbCOT} and \textit{LMLP} \cite{LMLP} leverage symbolic reasoning to validate or derive intermediate steps during inference. SymbCOT translates chain-of-thought (CoT) responses into first-order logic expressions, which are solved step by step by an LLM acting as a logic engine. A verification module then ensures semantic consistency and logic deduction validity. LMLP generalizes this idea to logic programs over a knowledge base (KB), verifying intermediate conclusions in knowledge-intensive QA tasks.

\textit{START} \cite{li2025startselftaughtreasonertools} builds a reasoning agent that learns to invoke symbolic tools adaptively. Hints are inserted into solution traces to prompt tool use at key reasoning junctures, and the model is fine-tuned via self-distillation. At inference time, the verifier strategically uses symbolic tools during sequential test-time scaling. 

\noindent \textbf{Symbolic Supervision for Training Verifiers:}
\textit{Deductive PRM} \cite{deductive_prm} proposes using symbolic representations, called \textit{natural programs}, which specify premises, conditions, and conclusions to guide the training of PRMs. These programs generated using strong LLMs (e.g., GPT-3.5-turbo), along with verification labels are distilled into smaller models like Vicuna through supervised fine-tuning (SFT).

\textit{FOVER} \cite{FOVER} uses formal verification tools such as Z3 and Isabelle to generate step-level annotations. Instead of verifying full proofs, the tools validate individual steps from reasoning traces generated by LLMs. This supervision signal is then used to train PRMs. Notably, FOVER achieves strong OOD generalization on reasoning benchmarks like ANLI, HANS, and MMLU.

\noindent\textbf{Neuro-Symbolic Self-Training}:
\textit{EnVISIONS} \cite{ENVISIONS} departs from static annotation paradigms by interacting with an external symbolic environment. The model outputs symbolic representations (e.g., first-order logic), which are executed and scored by the environment. The resulting binary feedback is converted into soft rewards and used for iterative refinement and self-training, enabling verification without human supervision.


\section{Benchmarks and Evaluations}
\label{sec:benchmarks}

We discuss different reference-based and reference-free benchmarks specifically created to evaluate process-based verifiers \cite{lightman2024lets,zheng2025processbenchidentifyingprocesserrors} or outcome based ones \cite{rm_bench,lambert2024rewardbenchevaluatingrewardmodels}. The benchmark design can also be mapped to \textit{asymmetry of verification} based on the tasks represented. When constructing these benchmarks, researchers generate multiple responses for the same query. During the manual annotation or automated judgment phase, for outcome-based verifier benchmarks, annotators are required to compare final responses or intermediate ones in case of process-level benchmarks and assign scores based on preference, correctness.

\textbf{RM-Bench} \cite{rm_bench} primarily focuses on  verifying ORM models and comprises tasks from domains code, math, open-ended chat and safety. Code and math are objective and hence are \textit{easy to verify} through unit tests and final answers, but verifying factual inaccuracies in open-ended chat and assessing whether responses are safe or not are a bit subjective and \textit{hard to verify}.

However, the existing ORM benchmarks primarily focus on preference-based evaluation, where verifiers are ranked based on their ability to compare competing solutions.


\textbf{Verifybench} \cite{yan2025verifybenchbenchmarkingreferencebasedreward} bridges this gap by curating a benchmark for reference-based reward models by focusing on objective accurate judgments with respect to a reference solution when compared to pairwise assessments aligning well with real-world applications of verification models such as test-time scaling or training of reasoning models. While this benchmark focuses on moderately easy to verify tasks, authors also introduce Verifybench-Hard which comprises of \textbf{hard to verify} instances

\textbf{PRM800K} \cite{lightman2024lets} primarily focused on collecting large volume of step-by-step reasoning based verification labels for math problems to benchmark PRMs. But it lacks task diversity and mostly comprises of easy to verify math problems. 
\textbf{ProcessBench} \cite{zheng2025processbenchidentifyingprocesserrors} proposed to  Olympiad level difficulty problems, with high task diversity. Unlike existing synthetic benchmarks, Processbench collected fine-grained human feedback for step-level annotations. While math domain is generally considered \textit{easy to verify}, Olympiad problems are more subjective to evaluate as verification does not only pertain to correctness of final answer but also creativity of the solution rendering it a \textit{hard to verify} task.

\noindent \textbf{Evaluation Metrics}:
Most of the benchmarks verify the correctness of error identification by the verifiers. The ORM benchmarks namely RM-Bench, Rewardbench employ accuracy of verifying the final output with respect to reference annotations as metric. Processbench computes step-level verification accuracy (earliest error detection) with respect to ground truth and employs F1 score to balance between being overly
critical and being incapable of identifying errors. PRM800K uses the final task performance as metric to determine quality of the verification rather than the accuracy of verification process itself.   
Additional details are in \textbf{Appendix \ref{app:benchmarks} with benchmark classification in Table \ref{tab:benchmark_matrix}} and corresponding metrics in \textbf{Table \ref{tab:verifier_metrics}}.

\section{Challenges and Future Directions}

\textbf{Limited Modalities}: Current verifier approaches for test-time scaling primarily focus on textual modality. This is primarily due to significant advances in test-time scaling being currently limited to text modality \cite{zhang2025surveytesttimescalinglarge, chung2025revisitingtesttimescalingsurvey}. However, though more recently scaling approaches for visual modality have emerged, corresponding verification approaches has not advanced significantly. While there is some initial research in this area namely VisualPRM \cite{wang2025visualprmeffectiveprocessreward}, a future direction could be to focus on developing principled verification methods beyond text.

\textbf{Efficiency of verifiers}:
A large number of verifier approaches primarily train a LLM or a LRM as ORM or PRM as outlined in this survey in Section \ref{sec:methods}. However, this requires additional compute compounding over compute used for solution candidates generation using another LLM with test-time scaling. Hence, there is a need for efficient verification approaches. While pruning and quantization are possible approaches for reducing memory requirements, they may lead to sub-par performance. One of the interesting future directions would be to develop novel learning approaches for verifiers employing Small Language Models (SLMs). Alternatively algorithms for assembling an ensemble of expert SLMs which is dynamically determined for a task so they can achieve robustness and generalization capabilities similar to their larger counterparts is an interesting direction. Data efficient training mechanisms and data augmentation for improving data quality are also possible directions that could improve performance of SLMs.

 \textbf{Generalization gap and Limited Benchmarks}: Many of the current state-of-the-art verifier training approaches still suffer from the out of distribution generalization gap \cite{sample_scrutinize_scale}. This also primarily stems from lack of benchmarks representative of diverse real-world tasks, as current benchmarks are majorly focused on code or math. While multi-domain data augmentation strategies \cite{zeng2025versaprm} have shown some promise, curating data for diverse tasks with manual annotations is time-consuming. Further augmenting them with step-level annotations manually for training verifiers becomes infeasible. While existing data augmentation approaches for training verifiers focus on generating step-level or final accuracy annotations \cite{zeng2025versaprm, khalifa2025processrewardmodelsthink,wang-etal-2024-math,luo2024improvemathematicalreasoninglanguage, autocv}, they still rely on all the queries and corresponding ground truth to be curated through manual annotation. This limits the query and task distribution that the verifier can observe limiting generalization.

 \textbf{Future Directions}: We envision a generalized framework for training verifiers that would contain the following characteristics: 1) generation of synthetic data with automated annotations for training outcome or process level verifiers 2) produce verifiers for wide-range of natural language reasoning tasks beyond math and code  and 3) produce verifiers that ideally help generalize out of distribution

To achieve 2) and 3) it is necessary for the verifier to be trained on diverse tasks and diverse queries corresponding to these tasks. Hence, 1) involves a generating queries for diverse tasks going beyond human annotated task distributions. The recent advances in reinforced self-play \cite{zhao2025absolutezeroreinforcedselfplay}, and verification based self-certainty based signals \cite{zhao2025learningreasonexternalrewards}, could lead to an unified framework comprising task proposer, task solver and verifier that co-evolve using specific rewards such as task diversity, step-wise correctness designed for each component. This coupled with signals from uncertainty quantification could be used as step-wise reward (without annotations) for training verifiers beyond math and code.

\vspace{-0.6em}
\section{Conclusion}
This work presents a survey of verification approaches scoped to the different types and training mechanisms for verifiers when employed for for test-time scaling. We define a clear taxonomy primarily focusing verifier types , training mechanisms, objectives and categorize the literature based on this taxonomy (Table \ref{tab:verification}). Finally, we highlight challenges in current verifiers and future directions.

\section{Limitations}

While our survey provides a comprehensive overview of verifier mechanisms that aid in test-time scaling, due to the rapid progress in the field, several adjacent works in reward models or reward design may have been overlooked. While reward design for pre-training is an interesting topic, our survey is scoped to reward models /verifiers for test-time scaling as in this setup verifiers guide search over vast candidate space. We will maintain this survey in form of repository to enable addition of new literature as the field progresses. Additionally, we do not focus on different prompt based reasoning mechanisms that briefly touch upon self-verification by asking LLM. While we have briefly discussed some of the important works in this area, adjacent works which do not primarily contribute to new approaches for verification or works where verification is not one of the core focus areas are omitted. This is primarily because fine-tuning based or symbolic approaches have shown to be more robust and calibrated than simply asking LLMs. Covering robust mechanisms for training or employing verifiers is one of the important aspects of our survey.


\section{Ethical Considerations and Risks}

The reliance on LLM-generated reasoning paths in test-time scaling introduces potential risks of reinforcing model biases or factual errors during verifier training with automatically generated data. Although the verifier is trained to select accurate reasoning, it may inherit systematic flaws from the base LLM. However, in our survey we also particularly focus on training mechanisms and verification approaches that help improve robustness of verifiers  which would help discard solution candidates with factual errors. However, the problem of hallucination in LLMs that lead to such issues is far from solved and building more robust verification mechanisms could help mitigate this issue.

Second, current landscape of verification approaches do not currently include fairness or bias mitigation across different attributes for tasks like Question Answering. While this is not the scope of this work, we believe this could be an important direction to build trustworthy systems.

\bibliography{references}
\bibliographystyle{acl_natbib}

\clearpage
\appendix

\begin{table*}[ht]
\centering
\small
\renewcommand{\arraystretch}{1.4}
\begin{tabular}{l|p{6.2cm}|p{6.2cm}}
\toprule
\textbf{Verification Type} & \textbf{Easy to Verify} & \textbf{Hard to Verify} \\
\midrule
\textbf{Outcome-level (ORM)} 
& \checkmark\ \textbf{RM-Bench} (code, math) -- Objective correctness via unit tests or exact numeric answers. 
\newline \checkmark\ \textbf{VerifyBench} -- Objective reference-based evaluation against gold answers. 
& \checkmark\ \textbf{RM-Bench} (safety, open-ended QA) -- Requires subjective judgment of safety and factual accuracy. 
\newline \checkmark\ \textbf{VerifyBench-Hard} -- Contains ambiguous references and subtle factual inconsistencies. \\
\midrule
\textbf{Process-level (PRM)} 
& \checkmark\ \textbf{PRM800K} -- Step-by-step math reasoning with deterministic correctness; easy to check. 
& \checkmark\ \textbf{ProcessBench} -- Olympiad-level math and creative proofs; requires subjective evaluation of reasoning path and novelty. \\
\bottomrule
\end{tabular}
\caption{Matrix mapping benchmarks to verification type (rows) and verification difficulty (columns), with reasons for categorization. Easy-to-verify tasks are typically objective (unit tests, deterministic solutions), whereas hard-to-verify tasks require subjective or nuanced evaluation (safety, creativity, subtle factuality).}
\label{tab:benchmark_matrix}
\end{table*}

\begin{table*}[ht]
\centering
\small
\renewcommand{\arraystretch}{1.3}
\begin{tabular}{p{3.5cm} p{6cm} p{6cm}}
\hline
\textbf{Benchmark} & \textbf{Metrics} & \textbf{ Rationale} \\
\hline
\textbf{PRM800K} & 
\begin{itemize}[leftmargin=*]
\item Step-level classification accuracy (correct/incorrect step labeling)
\item Downstream final-answer \texttt{pass@1} / \texttt{pass@k} improvement when PRM guides inference
\end{itemize}
& Provides large-scale human-labeled step correctness for math. PRMs trained on it can re-rank reasoning paths during inference, improving final task accuracy. \\
\hline 
\textbf{ProcessBench} & 
\begin{itemize}[leftmargin=*]
\item Earliest-error detection accuracy
\item F1 score for identifying first incorrect step
\item Final-task accuracy improvement when verifier guides inference
\end{itemize}
& Focuses on localizing the exact failure point in reasoning chains. F1 balances false positives/negatives. Evaluates both detection quality and downstream accuracy gains when used in test-time scaling. \\

\hline

\textbf{VerifyBench \& VerifyBench-Hard} & 
\begin{itemize}[leftmargin=*]
\item Accuracy of binary correct/incorrect judgment vs. reference answer
\end{itemize}
& Evaluates reference-based correctness judgments. The ``Hard'' split contains subtle or ambiguous examples to test robustness. \\
\hline 

\textbf{RM-Bench} & 
\begin{itemize}[leftmargin=*]
\item Pairwise accuracy (chosen $>$ rejected)
\item Accuracy breakdown by difficulty (easy, medium, hard)
\end{itemize}
& Evaluates how well outcome reward models rank human-preferred completions over rejected ones, with breakdown by difficulty. \\
\hline 
\textbf{RewardBench} & 
\begin{itemize}[leftmargin=*]
\item Pairwise accuracy across prompt--chosen--rejected triples
\item Domain-specific accuracy breakdown (safety, reasoning, chat, etc.)
\end{itemize}
& Measures preference-based ranking performance across diverse domains; tests cross-domain generalization of reward models. \\
\hline
\end{tabular}
\caption{Evaluation metrics for major verifier benchmarks used in test-time scaling.}
\label{tab:verifier_metrics}
\end{table*}

\section{Literature Compilation}
\label{app:literature_compilation}
\subsection{Search Strategy}

We conducted a comprehensive search on Google Scholar. We first focused on highly relevant Natural language Processing (NLP) venues such as ACL, EMNLP, NAACL, COLM and journals like TACL. Since, a lot of verifier advancements come from the ML community we also focused on venues like NeurIPS, ICLR, AAAI and ICML. Since this area is rapidly moving we also focused on arXiV pre-prints. We employed keywords like  ``verifiers", ``test-time scaling", ``self-verification", ``process reward models", ``outcome reward models" to retrieve relevant surveys and works. We focus on papers till march 2025 and also include some recent paper in July 2025. however, owing to rapid progress in this topic, some latest pre-prints could have been excluded inadvertently from this survey.
\subsection{Selection Strategy}

We primarily focus on verification approaches that help verify the final answer (ORM) or step-by-step reasoning process for arriving at solution (PRM) in context of test-time scaling. While reward models which also help value outputs during pre-training have similar objectives they are always not used in guiding exploration or search for solutions. Hence, they are excluded from our survey as our focus is on verification approaches for test-time scaling. this filtering was done based on careful review of abstract, Introduction, Conclusion and Limitations sections of these papers. We also  focused on the Methodology section of these papers to compile papers that detailed different training mechanisms for verifiers which is one fo the key focus areas of our survey. After careful review, 42 papers (Table \ref{tab:verification}) were selected which forms the foundational content of this survey.

\section{Additional Benchmark and Evaluation Details}

Expanding upon benchmarks discussed in Section \ref{sec:benchmarks}, we expand upon the statistics of these benchmarks in this section.
An overview of benchmarks and their categorization based on verification types and asymmetry of verification are as shown in Table \ref{tab:benchmark_matrix} providing a mental model of the discussion in Section \ref{sec:benchmarks}.

\label{app:benchmarks}

\textbf{PRM800K} is among the first large-scale, human-labeled datasets with fine-grained feedback on reasoning steps. The final dataset comprises 800,000 step-level labels across ~75,000 solution traces from ~12,000 problems.

The PRM800K dataset was curated in two major phases:

\begin{itemize}
    \item \textbf{Phase 1} (\(\sim 5\%\) of data): Randomly sampled solution traces generated by an LLM were labeled by human annotators.
    \item \textbf{Phase 2} (\(\sim 95\%\) of data): An \textit{active learning} strategy was adopted:
    \begin{enumerate}
        \item Generate multiple candidate solutions per problem.
        \item Use the current PRM to rank the solutions.
        \item Present the most \emph{convincing incorrect} answers to human annotators for labeling.
    \end{enumerate}
    This approach reduces redundant labeling and prioritizes informative, borderline cases.
\end{itemize}

\textbf{Labeling Scheme}:

Each reasoning step in a solution trace is annotated with one of three labels:
\begin{itemize}
    \item \textbf{Positive (+)}: Step is correct and helpful.
    \item \textbf{Neutral (0)}: Step is ambiguous or correct but non-progressive.
    \item \textbf{Negative (--)}: Step contains an error or irrelevant content.
\end{itemize}

\textbf{RM-Bench} It comprises  approximately 2,000–3,000 preference instances, across multiple domains: Chat, Code, Math, Safety. It was created to evaluate outcome reward models' sensitivity to subtle content changes and robustness against style bias, which traditional benchmarks overlooked for open-ended chat tasks. it also comprises objective tasks like code and math which are easy to verify.

\textbf{Verifybench} was primarily curated to benchmark reference-based verifiers. It  comprises 2,000 well-balanced question-answer-completion-correctness tuples. Instructions and gold reference answers were sourced from established open datasets across diverse reasoning domains (e.g., GSM8K, MATH500, MultiArith, ProofWriter). For each instruction, multiple candidate completions were generated using both open-source and proprietary LLMs (e.g., GPT‑4o, Qwen) to cover a spectrum of success levels. Each completion is labeled by human annotators as correct or incorrect based on alignment with the reference answer, without preference comparisons. Verification is binary and absolute, not relative which requires that classifiers predict true/false alignment with ground truth

\begin{table*}[hbt!]
    \centering
    \vspace{{-2em}}
    \resizebox{1\textwidth}{!}{
    \begin{tabular}{llp{7.8cm}p{5cm}}
    \toprule
        \rowcolor{gray!10}
        \textbf{Category} & \textbf{Approach} & \textbf{Hypothesis} & \textbf{Training Recipe}  \\
    \midrule
        \multirow{3}{*}{\textbf{Heuristic}}     & (1) Deepseek-R1~\citep{deepseek-r1} &  &  - \\

        & (2) RvLLM \cite{rvllm_rule_based} & Domain specific rules for verification leads to more deterministic outcomes & - \\

           \hline
     


       \textbf{Generative} \\
       - Classical SFT & (3) Generative Verifier~\citep{zhang2024generative} &  &  \\

        & (4) SyncPL \cite{SyncPL} & Trains a generative process verifier using criteria based data generation & Employs SFT to fine-tune a LLM to generate critiques and correctness label for solution candidates \\
        \cmidrule{2-4}

        & (5) UQ-PRM \cite{ye2025uncertaintyaware} & Uncertainty quantification of Process verifiers leads to more robust reasoning in LLMs & Stepwise Uncertainty Quantification is included as part of training process for verifiers.  \\
        \cmidrule{2-4}

        & (6)  WoT~\citep{zhang-etal-2024-wrong} & Verification process is not uni-dimensional and & Scaling verifiers \\

        & (7) Multi-Agent Verifiers~\citep{lifshitz2025multiagent} &  hence the verifier must cover diverse perspectives of this process & to cover diverse task-related  aspects that need to be verifier  \\ 

        \midrule
        - Self-verification  & & & \\
        (classical SFT) 
        & (8) ReVISE \cite{lee2025revise} & Traiining to self-verify through synthetic data  & Training using imitation learning on  \\
        & (9) Self-Reflection Feedback~\citep{li2025learningreasonfeedbacktesttime} & and external feedback leads to iterative self-improvement and efficient verification  & iterative self-reflection based synthetic data \\
        \cmidrule{2-4}
        & (10) ToolVerifier \cite{mekala-etal-2024-toolverifier} & Self-verification through tool use  &  \\
        & (11) T1 \cite{t1} &  or data from tool use leads to more robust verification. & SFT of LLM on data from tool use  \\

            
        \cmidrule{2-4}
       - RL-based 
        & (12) GenRM \cite{mahan2024generativerewardmodels} & Verifier training through preference optimization & Trains generative verifiers using preference optimization \\
       
       &(13)  V-STaR~\citep{hosseini2024vstar}  & using automatically curated data or from human feedback offers better alignment to preferred responses. &  like Direct Preference Optimization (DPO) \\
       \cmidrule{2-4}


       & (14)  RL-Tango \cite{zha2025rltangoreinforcinggenerator} & Jointly training Generator and Verifier together  & RL-based Co-evolution \\

       & (15)  S2R \cite{ma-etal-2025-s2r} & is better than training a verifier with fixed generator & of Generator and Verifier \\

\midrule
         {\textbf{Reasoning}} & (16) Heimdall \cite{heimdall_reasoning_verification} & Training a reasoning model & Trains a Large Reasoning Model \\
        & (17) Think-J \cite{thinkJ} & on joint objective of reasoning or critique  & on generative and solution   \\

        & (18) ThinkPRM \cite{khalifa2025processrewardmodelsthink} & generation and correctness prediction& correctness prediction tasks to . \\
       & (19) J1 \cite{whitehouse2025j1incentivizingthinkingllmasajudge}& renders it an effective judge for verification tasks & serve as judge \\
       \cmidrule{2-4}
        & (20) $RL^V$ \cite{rlv_reasoning} & Scaling generative verifier training by employing    & Trains a LRM on generative verification and RL objectives jointly  \\
        & (21) OREO \cite{OREO} & reasoning models with RL objectives enables bettwe exploration of solution space than SFT based imitation learning  & Trains a value function by optimizing soft Bellman equation for test-time scaling\\
  \cmidrule{2-4}
              - Adaptive & (22) DyVe \cite{zhong2025dyvethinkingfastslow} & Training an adaptive verifier through imitation   & SFT on curated data  \\
        & (23) FlexiVe \cite{flexive} &  learning (SFT) or RL to alternate between fast \& slow thinking leads to efficient TTS & RL based adaptive verifier.  \\
    \midrule


      \textbf{Symbolic}    
      & (24) RaLU \cite{li2025reasoningaslogicunitsscalingtesttimereasoning} & Symbolic reasoning through program generation & Trains to generate accurate \\

      & (25) SymbCOT \cite{symbCOT} & aids in more structured LLM-based verification.  & symbolic programs to aid in verification \\
  \cmidrule{2-4}

        & (26) START~\citep{li2025startselftaughtreasonertools} & Self-distillation for learning to use symbolic tools  & Trains a reasoning agent to invoke symbolic tools adaptively using instruction tuning through self-ditillation.  \\

       &  (27) LMLP \cite{LMLP} &  for step-wise verification improves factuality of generated responses & Trains to use programs that can be run over KB for step-wise verification.
        \\
        \cmidrule{2-4}
        & (28) ENVISIONS \cite{ENVISIONS} & Neruo-symbolic self-training than distilling from stronger LLMs helps overcome the scarcity of symbolic data, and improves the proficiency of LLMs in processing symbolic language & Self-training framework for verifiers using interactive feedback from symbolic environment than static annotations \\

  \cmidrule{2-4}
&  (29) Deductive PRM~\citep{deductive_prm} & Verifier trained through Distilling symbolic program generation capabilities  & Symbolic supervision for training verifiers \\
& (30) FOVER \cite{FOVER} & \& verification labels from formal verification tools leads to deterministic and accurate verification  & 
 \\       
 
  
    \bottomrule
    \end{tabular}}
    \vspace{-0.8em}
    \caption{Summary of verifier training mechanisms as applied to test-time scaling. KB- Knowledge Base, }
    \label{tab:hypothesis}
\end{table*}

\textbf{ProcessBench} consists of 3,400 test cases which are high quality Olympiad math problems, with all solutions annotated with step-wise labels indicating which steps are accurate or inaccurate by multiple human experts. The expert annotation ensure the
data quality and the reliability of evaluation. The annotators were instructed to identify the following types of errors for step-wise reasoning for constructing ProcessBench:
(1) Mathematical errors: incorrect calculations, algebraic manipulations, or formula applications. (2)
Logical errors: invalid deductions, unwarranted
assumptions, or flawed reasoning steps. (3) Conceptual errors: misunderstanding or misapplication of mathematical or problem concepts. (4)
Completeness errors: missing crucial conditions,
constraints, or necessary justifications that affect
the solution’s validity.

\noindent \textbf{RewardBench} comprises of instances from different task categories namely open-ended chat (easy and hard), Reasoning (code and math) and a task which tests for safety of LLMs. For open-ended chat prompts, preferred and rejected response pairs are chosen from Alpaca Eval \cite{alpaca_eval} and 70 instance from MT-bench \cite{zheng2023judgingllmasajudgemtbenchchatbot}

For the safety category, the goal is to test models’ tendencies to refuse dangerous content and to avoid incorrect refusals to similar trigger words. Prompts and chosen, rejected pairs are selected from custom versions of the datasets XSTest \cite{rottger-etal-2024-xstest} and Do-Not-Answer \cite{wang-etal-2024-answer}.

\noindent \textbf{Evaluation metrics}
Expanding upon metrics discussed in Section \ref{sec:benchmarks}, Table \ref{tab:verifier_metrics} provides a detailed overview of evaluation metrics employed for verifiers on different benchmarks with the corresponding explanation (rationale).

\end{document}